\documentclass{article}
\usepackage{enumitem}
\usepackage{spconf,amsmath,graphicx}
\usepackage{multirow}
\usepackage{xcolor}
\usepackage{subcaption}
\usepackage[version=4]{mhchem}
\usepackage{amsfonts}
\usepackage{booktabs}
\usepackage{cite}

\usepackage[bookmarks=true,breaklinks=true,colorlinks,linkcolor=blue,citecolor=blue,urlcolor=blue]{hyperref}
\usepackage{comment}
\usepackage{xspace}

\newcommand{\design}{{VQEzy}\xspace}

\title{
\textnormal{
\design: An Open-Source Dataset for Parameter Initialization in \\Variational Quantum Eigensolvers
\vspace{-0.1in}}
}

\name{}
\address{}

\name{Chi Zhang$^*$ \quad Mengxin Zheng$^\dagger$ \quad Qian Lou$^\dagger$  \quad Hui Min Leung$^*$ \quad Fan Chen$^*$\vspace{-12pt}}
\address{$^\dagger$University of Central Florida, FL, USA\\
$^*$Indiana University, Bloomington, IN, USA\vspace{-8pt}}

\renewenvironment{thebibliography}[1]{%
   \begin{oldthebibliography}{#1}%
     \setlength{\itemsep}{-.4ex}%
}%
{%
   \end{oldthebibliography}%
}

\begin{document}
\maketitle

\begin{abstract}
Variational Quantum Eigensolvers (VQEs) are a leading class of noisy intermediate-scale quantum (NISQ) algorithms, whose performance is highly sensitive to parameter initialization. 
% Effective initialization improves trainability and reduces the risk of convergence to suboptimal local minima.
Although recent machine learning–based initialization methods have achieved state-of-the-art performance, their progress has been limited by the lack of comprehensive datasets. Existing resources are typically restricted to a single domain, contain only a few hundred instances, and lack complete coverage of Hamiltonians, ansatz circuits, and optimization trajectories. To overcome these limitations, we introduce \design, the first large-scale dataset for VQE parameter initialization. \design spans three major domains and seven representative tasks, comprising 12,110 instances with full VQE specifications and complete optimization trajectories. The dataset is available at \href{https://github.com/chizhang24/VQEzy}{https://github.com/chizhang24/VQEzy}, and will be continuously refined and expanded to support future research in VQE optimization.
\end{abstract}

\vspace{-6pt}
\section{Introduction}
\label{sec:intro}
\vspace{-6pt}

\textbf{Related Work and Motivation}.  
The Variational Quantum Eigensolver (VQE)~\cite{peruzzo2014variational, mcclean2016theory} is a prominent algorithm for the NISQ era, with applications in many-body physics~\cite{bharti2022noisy, bauer2020quantum, hensgens2017quantum}, quantum chemistry~\cite{li2019variational}, and related fields~\cite{tilly2022variational}. A VQE framework comprises three key components: the Hamiltonian, a parameterized quantum circuit, and a classical optimizer. Despite its potential, VQE training remains challenging, as convergence is highly sensitive to parameter initialization. Effective initialization enhances trainability and mitigates convergence to suboptimal local minima.

Recent parameter initializers leverage machine learning models~\cite{mesman2024nn, Miao:PRA2024, zhang2025diffusion} trained on optimized VQE parameters to generate effective initializations for new tasks. 
However, their development are limited by the absence of comprehensive VQE datasets.  
Existing datasets suffer from three main limitations:  
(1) domain restriction, focusing only on physics~\cite{liQASMBenchLowLevelQuantum2023, nakayama2023vqegenerated,leeQMAMLQuantumModelAgnostic2025a} or chemistry~\cite{Utkarsh2023Chemistry};  
(2) small scale, typically a few hundred instances (e.g., fewer than 500 in~\cite{leeQMAMLQuantumModelAgnostic2025a}); and  
(3) incomplete coverage, with some containing only Hamiltonians and/or ansatz circuits~\cite{liQASMBenchLowLevelQuantum2023} or omitting ansatz data entirely~\cite{Utkarsh2023Chemistry}.

\begin{figure}[t!]
    \centering
    \includegraphics[width=1.0\linewidth]{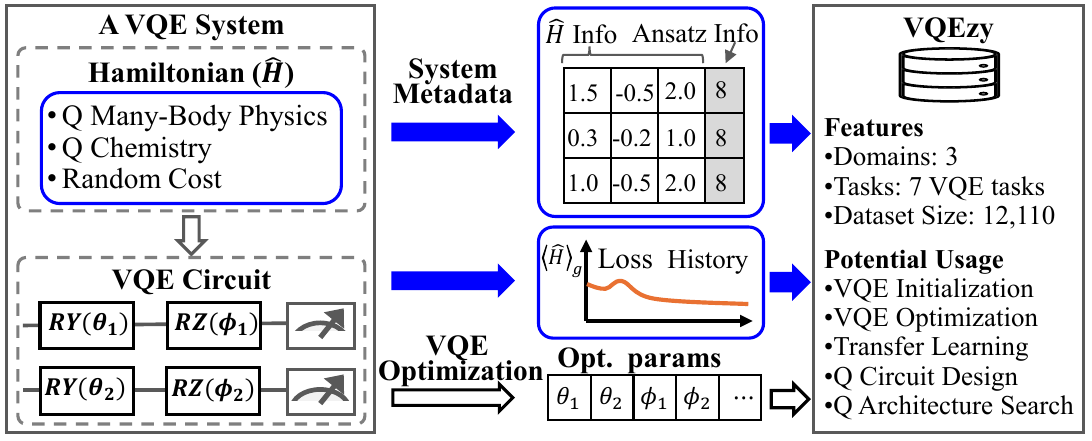}
    \vspace{-0.25in}
    \caption{Overview of the~\design~dataset. \textcolor{blue}{Blue} marks key advances over prior work: (1) coverage of three domains and seven tasks with 12,110 instances; (2) inclusion of full VQE specifications and complete optimization trajectories.}
    \label{fig:VQEzy_overview}
    \vspace{-0.25in}
\end{figure}

\begin{table*}[t!]
\footnotesize
\caption{Summary of the~\design~dataset. Abbreviations: 1D\_XYZ – 1D Heisenberg XYZ model; 1D\_FH – 1D Fermi–Hubbard model; 2D\_TFI – 2D Transverse-Field Ising model; Feature Dim. – Feature Dimension.}
\label{tab:dataset_summary}
\vspace{-6pt}
\centering
\setlength{\tabcolsep}{4.6pt}
\begin{tabular}{|c| c| c| c |c |c| c| c| c|}\toprule
    \multirow{2}{*}{\textbf{Application}} 
    & \multirow{2}{*}{\textbf{Task}}  
    &\multicolumn{2}{c|}{\textbf{VQE Setup}}     
    &\multicolumn{3}{c|}{\textbf{Data Attributes}} 
    & \multirow{2}{*}{\textbf{Feature Dim.}}
    & \multirow{2}{*}{\textbf{\# Instances}}\\\cline{3-7} 
    
    & 
    & \textbf{Hamiltonian}
    & \textbf{Ansatz}
    & \textbf{Hamiltonian Parameters} 
    & \textbf{\# Params}  
    & \textbf{\# Qubits}  
    & & \\\toprule
    
    \multirow{6}{*}{\parbox{0.66in}{Quantum Many-Body Physics}} 
    & \multirow{2}{*}{1D\_XYZ}  
    & \multirow{2}{*}{Eq.~\ref{eq:heisenberg_xyz}} & \multirow{6}{*}{Fig.~\ref{fig:ansatz}(a)} & \multirow{2}{*}{Coupling constants $(J_1, J_2, J_3)$} & 8 & 4 & 1$\times$8 & 2000\\
    & &  & & & 24 & 12 & 1$\times$24 & 2000\\\cline{2-3}\cline{5-9}

     & \multirow{3}{*}{1D\_FH} & \multirow{3}{*}{Eq.~\ref{eq:1d_fh}} & & \multirow{3}{*}{Hopping $t$ and interaction $U$} & 8 & 4 & 1$\times$8 & 1000\\
    & &  & & & 12 & 6 & 1$\times$12 & 1000\\
    &  &  & & & 16 & 8 & 1$\times$16 & 1000\\\cline{2-3}\cline{5-9}
    
    & 2D\_TFI  & Eq.~\ref{eq:2d_tfi} &  & Coupling $j$ and field strength $\mu$ & 16 & 8 & 1$\times$16 & 1000\\\hline\hline

    \multirow{3}{*}{\parbox{0.66in}{Quantum Chemistry}} 
    & \ce{H2} & \cite{Utkarsh2023Chemistry} & \multirow{3}{*}{Fig.~\ref{fig:ansatz}(b)} & H–H bond length & 24 & 4 & 1$\times$24 & 150\\\cline{2-3}\cline{5-9}
    & \ce{HeH+} & \cite{Utkarsh2023Chemistry} & & He-H bond length & 24 & 4 & 1$\times$24 & 1000\\\cline{2-3}\cline{5-9}
    & \ce{NH3} & \cite{Utkarsh2023Chemistry} & & N-H bond length & 336 & 16 & 1$\times$336 & 160\\\hline\hline

    \multirow{1}{*}{\parbox{0.66in}{Random VQE}} 
    & Random\_VQE & \cite{liQASMBenchLowLevelQuantum2023} &Fig.~\ref{fig:ansatz}(c) & Pauli string coefficients & 48 & 4 & 1$\times$48 & 2800\\\bottomrule
\end{tabular}
\vspace{-0.05in}
\end{table*}
%%%%%%%%%%%%%%%%%%%%%%%
\begin{figure*}
   \centering
   \includegraphics[width=.98\linewidth]{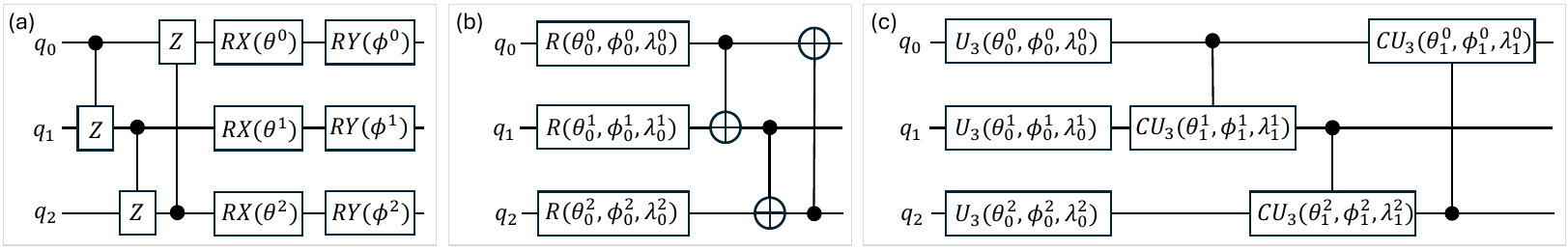}
    \vspace{-0.1in}
    \caption{Ansatzes used for (a) Quantum Many-Body Physics~\cite{zhangEscapingBarrenPlateau2022}, (b) Quantum Chemistry~\cite{schuldCircuitcentricQuantumClassifiers2020}, and (c) Random VQE~\cite{hanruiwang2022quantumnas}.}
   \label{fig:ansatz}
    \vspace{-0.15in}
\end{figure*}

\textbf{Contributions.}
This work addresses the above limitations by open-sourcing~\design, the first large-scale dataset for VQE parameter initialization. An overview of~\design~is shown in Figure~\ref{fig:VQEzy_overview}, and the key contributions are:

\begin{itemize}[leftmargin=*, topsep=-2pt, partopsep=-2pt, itemsep=-2pt]
    \item \textbf{Wide VQE Application Domains.} 
    \design~spans three main VQE domains—quantum many-body physics, quantum chemistry, and random benchmarking—with seven representative tasks and varied circuit implementations across qubit sizes and ansatz families. In total, \design~offers 12,110 VQE instances, orders of magnitude larger and richer than prior datasets~\cite{liQASMBenchLowLevelQuantum2023, nakayama2023vqegenerated,leeQMAMLQuantumModelAgnostic2025a}.

    \item \textbf{Comprehensive VQE Data Attributes.} 
    For each instance, we provide the optimized VQE parameter vector along with rich attributes—problem Hamiltonians, circuit specifications, and full optimization trajectories—supporting diverse research use cases.
   
    \item \textbf{Open-Source for Broad Access and Continuous Expansion.} 
    We open-source the \design~dataset and will continue adding more data points and features with the participation of the research community. We expect it to establish a foundation for VQE research and enable broader advances in VQE initialization, transfer learning across tasks, VQE architecture design, and beyond.
\end{itemize}

\vspace{-0.1in}
\section{\design Dataset}
\label{sec:dataset}
\vspace{-0.1in}

The overall approach for collecting optimized VQE initialization parameters consists of three stages: 
(1) problem Hamiltonian generation; 
(2) ansatz circuit selection; and 
(3) VQE optimization. 
This section details the construction of these stages in the proposed~\design~dataset.
A summary of the~\design~dataset is presented in Table~\ref{tab:dataset_summary}.

%%%%%%%%%%%%%%%%%%%%%%%%%%%%%%%%%%%%%%%%%%%%%%%%%%%%%%%%%%%%%%%%%%%%%
\begin{figure*}[t]
\centering
\makebox[\linewidth][c]{%
  \includegraphics[height=14pt]{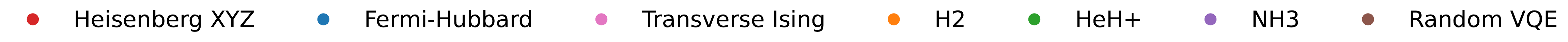}%
}
\vspace{1pt}
\makebox[\textwidth][c]{%
\begin{minipage}[t]{0.33\textwidth}
    \centering
    \includegraphics[width=0.5\textwidth]{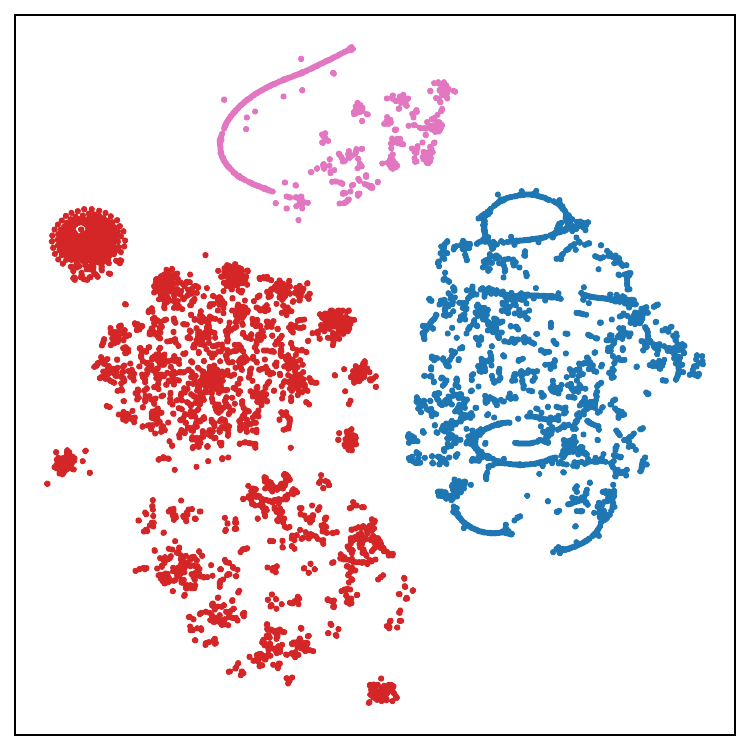}%
    \includegraphics[width=0.5\textwidth]{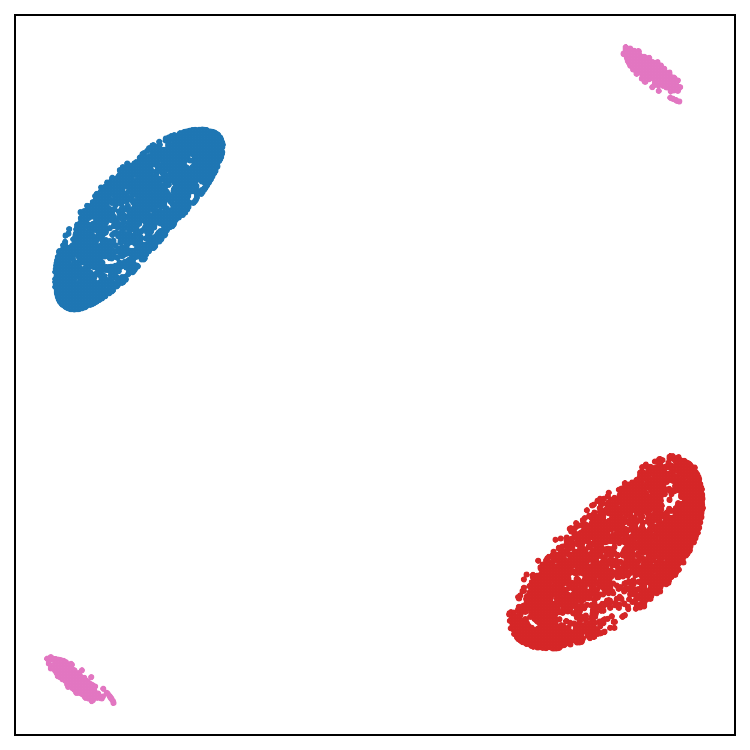}
    \vspace{0pt}
    {\small (a) Quantum Many-Body Physics}
\end{minipage}
\hspace{2pt}
%%%%%%%%%%%%%%%%%%%%%%%%%%%%%%%%
\begin{minipage}[t]{0.33\textwidth}
    \centering
    \includegraphics[width=0.5\textwidth]{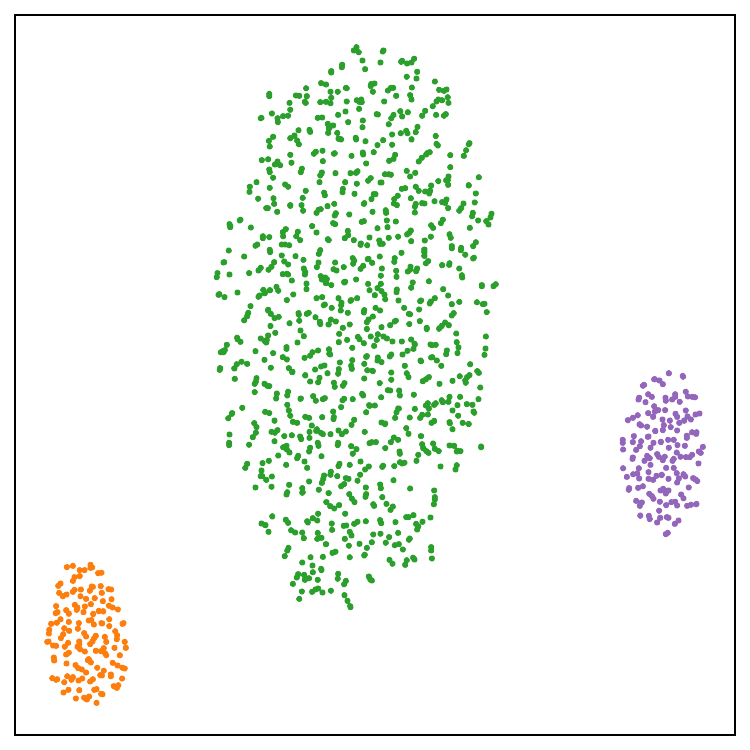}%
    \includegraphics[width=0.5\textwidth]{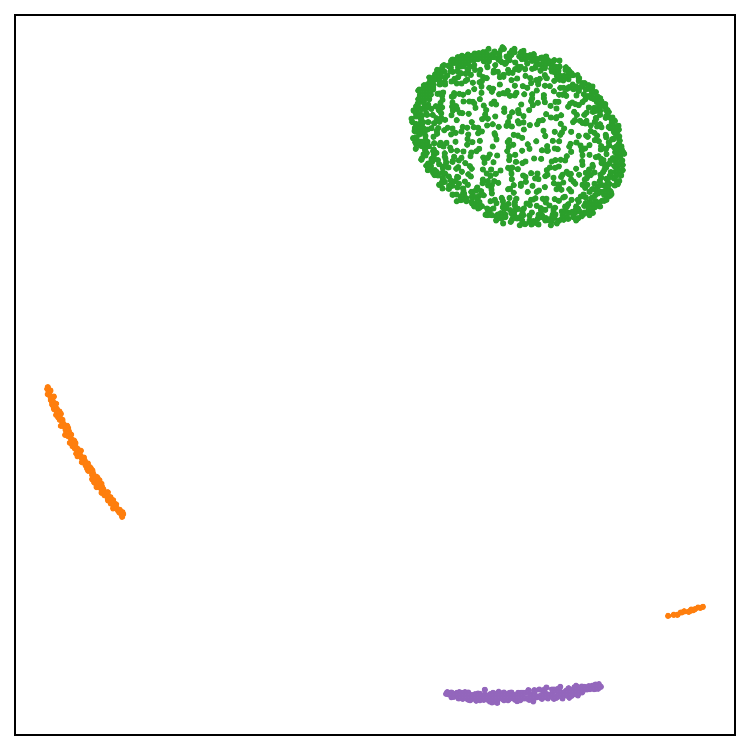}
    \vspace{0pt}
    {\small (b) Quantum Chemistry}
\end{minipage}
\hspace{2pt}
%%%%%%%%%%%%%%%%%%%%%%%%%%%%%%%%
\begin{minipage}[t]{0.33\textwidth}
    \centering
    \includegraphics[width=0.5\textwidth]{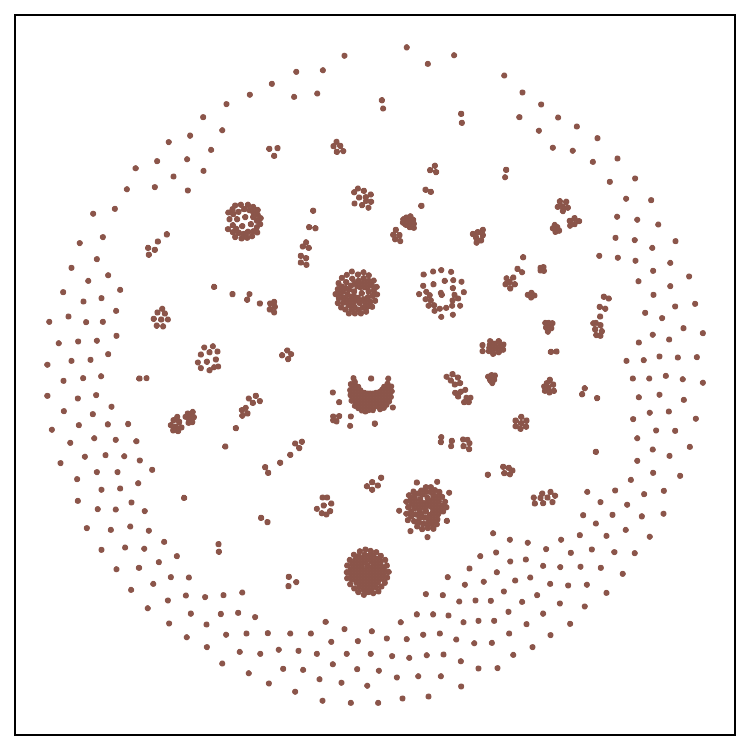}%
    \includegraphics[width=0.5\textwidth]{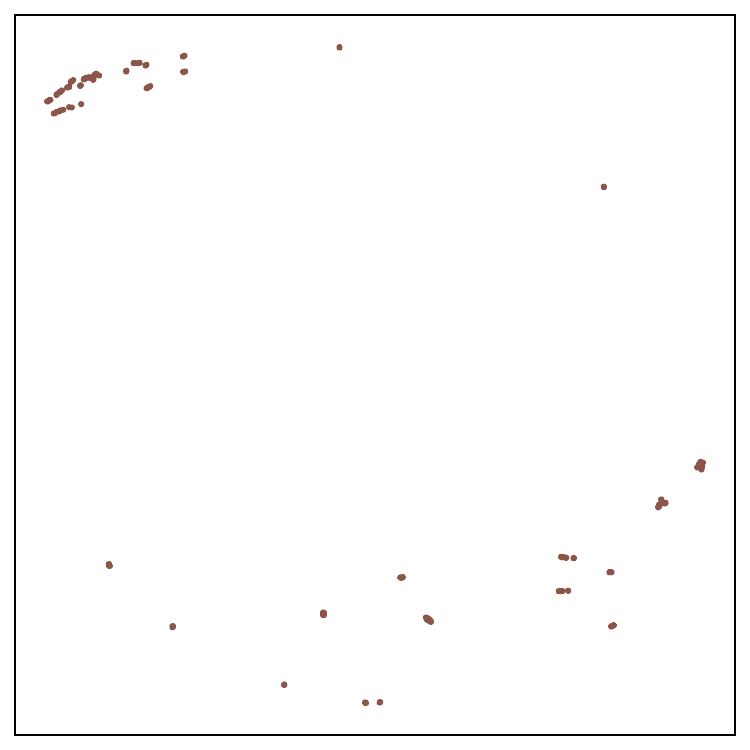}
    \vspace{0pt}
    {\small (c) Random VQE}
\end{minipage}
}
\vspace{-10pt}
\caption{Visualization of optimized VQE parameters in~\design~using t-SNE (left) and MDS (right) across application domains: (a) Quantum Many-Body Physics, (b) Quantum Chemistry, and (c) Random VQE.}
\label{fig:domain_tsne_mds}
\vspace{-0.12in}
\end{figure*}
%%%%%%%%%%%%%%%%%%%%%%%%%%%%%%%%%%%%%%%%%%%%%%%%%%%%%%%%%%%%%%%%%%%%%
\begin{figure*}[t]
\centering

\makebox[\textwidth][c]{%
\begin{minipage}[t]{0.33\textwidth}
    \centering
    \includegraphics[width=0.5\textwidth]{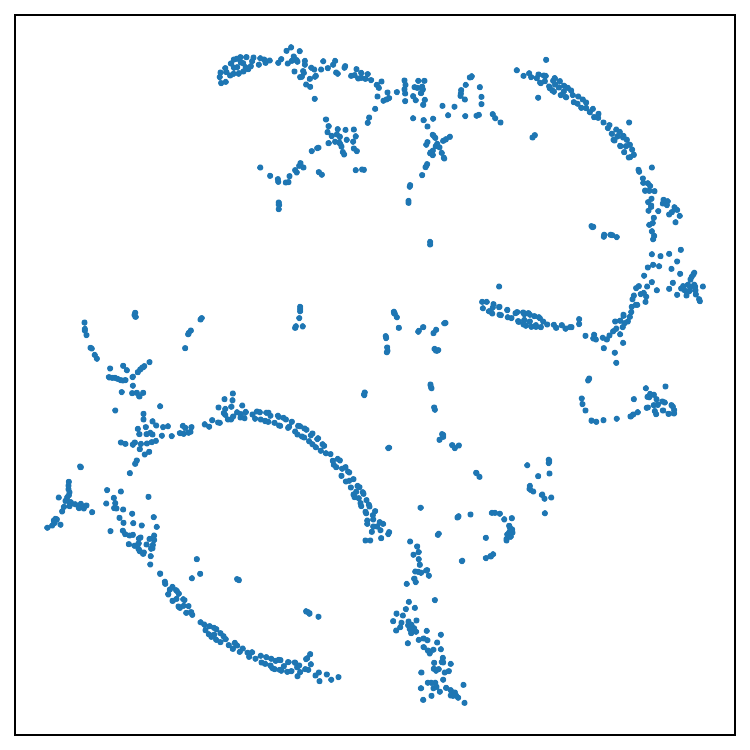}%
    \includegraphics[width=0.5\textwidth]{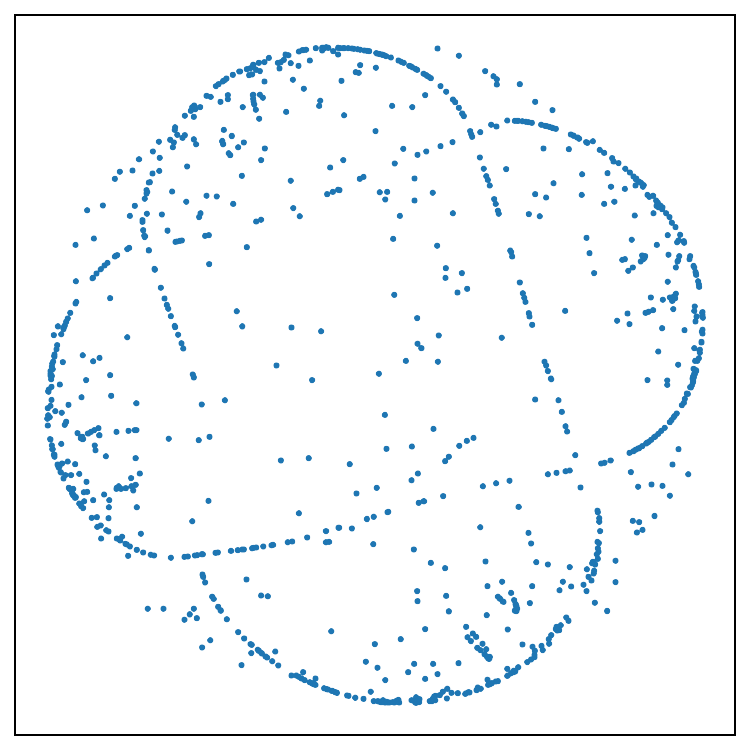}
    \vspace{0pt}
    {\small (a) 4-Qubit 1D\_FH}
\end{minipage}
\hspace{2pt}
%%%%%%%%%%%%%%%%%%%%%%%%%%%%%%%%
\begin{minipage}[t]{0.33\textwidth}
    \centering
    \includegraphics[width=0.5\textwidth]{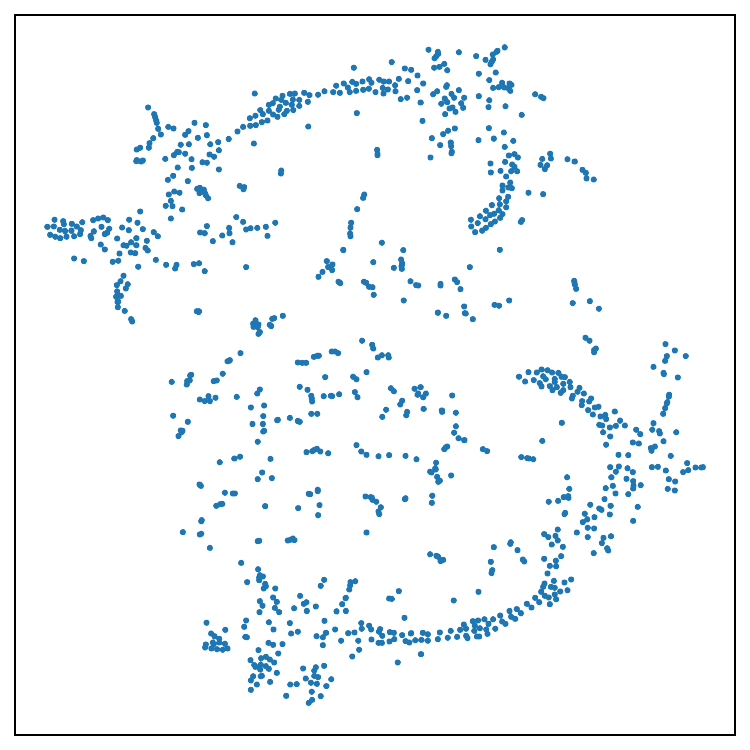}%
    \includegraphics[width=0.5\textwidth]{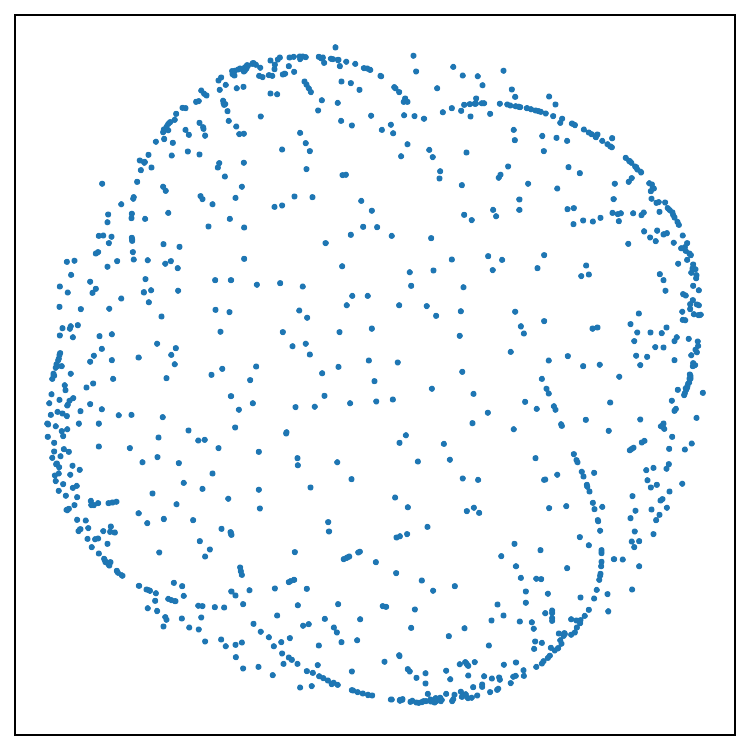}
    \vspace{0pt}
    {\small (b) 6-Qubit 1D\_FH}
\end{minipage}
\hspace{2pt}
%%%%%%%%%%%%%%%%%%%%%%%%%%%%%%%%
\begin{minipage}[t]{0.33\textwidth}
    \centering
    \includegraphics[width=0.5\textwidth]{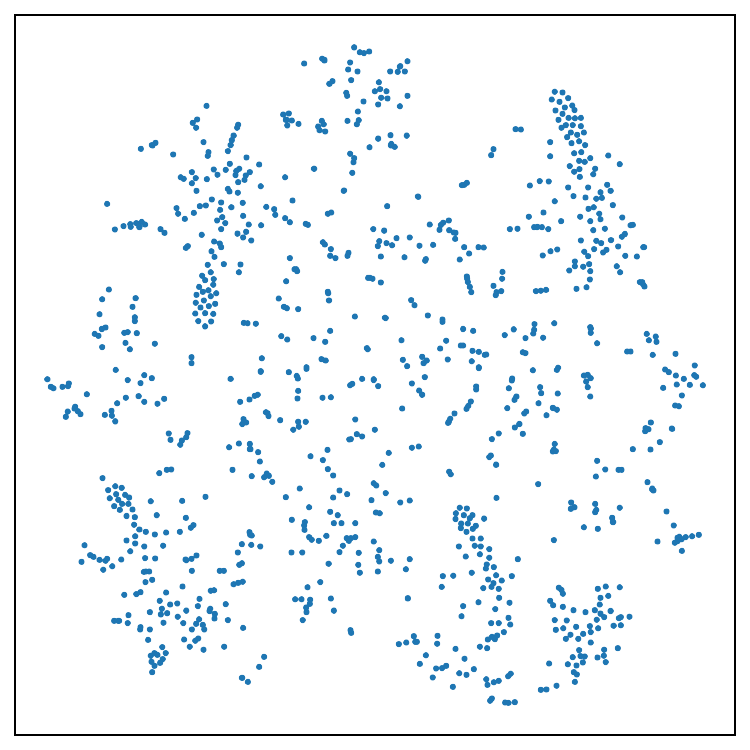}%
    \includegraphics[width=0.5\textwidth]{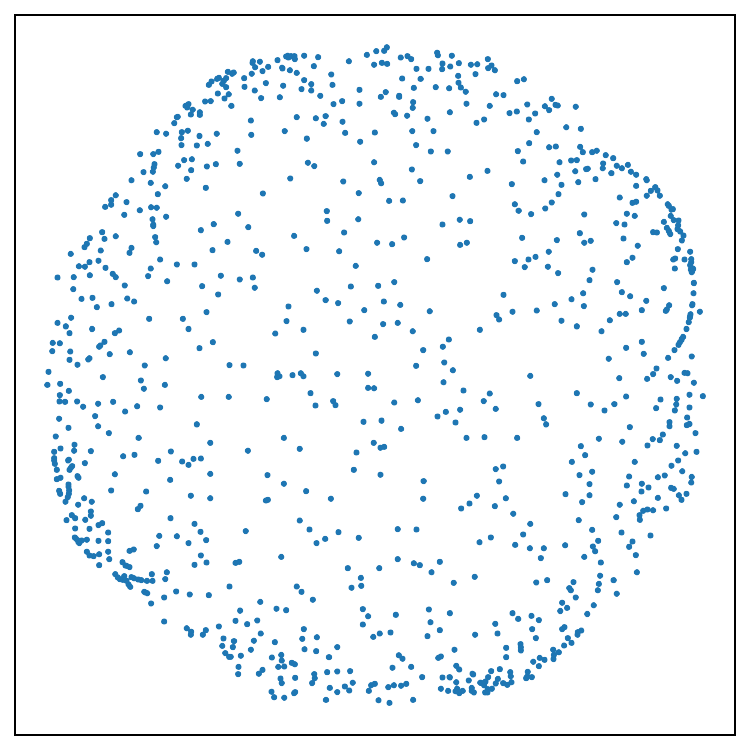}
    \vspace{0pt}
    {\small (c) 8-Qubit 1D\_FH}
\end{minipage}
}
\vspace{-8pt}
\caption{Visualization of optimized parameters using t-SNE (left) and MDS (right) for the 1D Fermi–Hubbard model with circuit implementations of (a) 4 qubits, (b) 6 qubits, and (c) 8 qubits.}
\label{fig:fh_tsne_mds}
\vspace{-0.22in}
\end{figure*}
%%%%%%%%%%%%%%%%%%%%%%%%%%%%%%%%%%%%%%%%%%%%%%%%%%%%%%%%%%%%%%%%%%%%%
\begin{figure}[t] % single-column figure
\centering

\begin{minipage}[t]{0.48\columnwidth}
    \centering
    \includegraphics[width=0.48\linewidth]{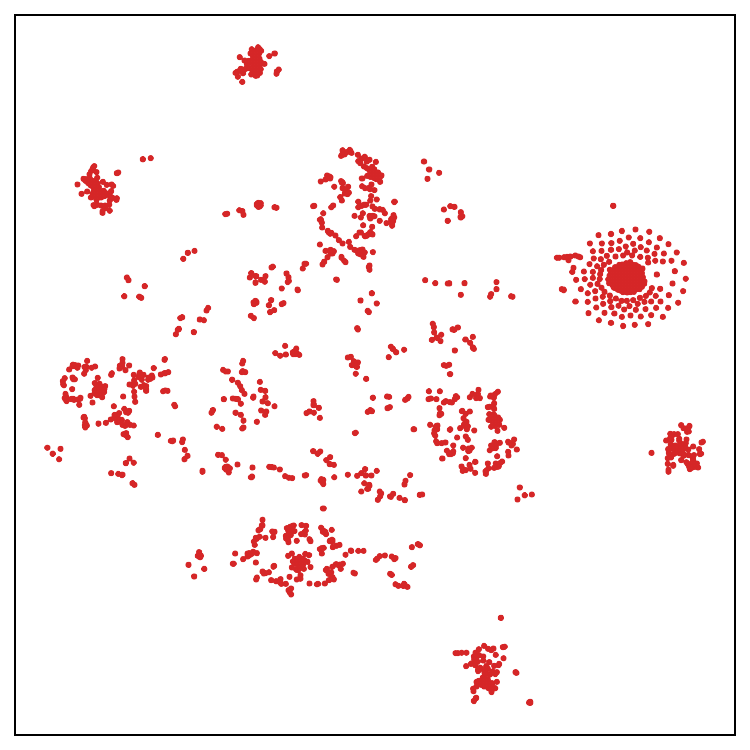}%
    \includegraphics[width=0.48\linewidth]{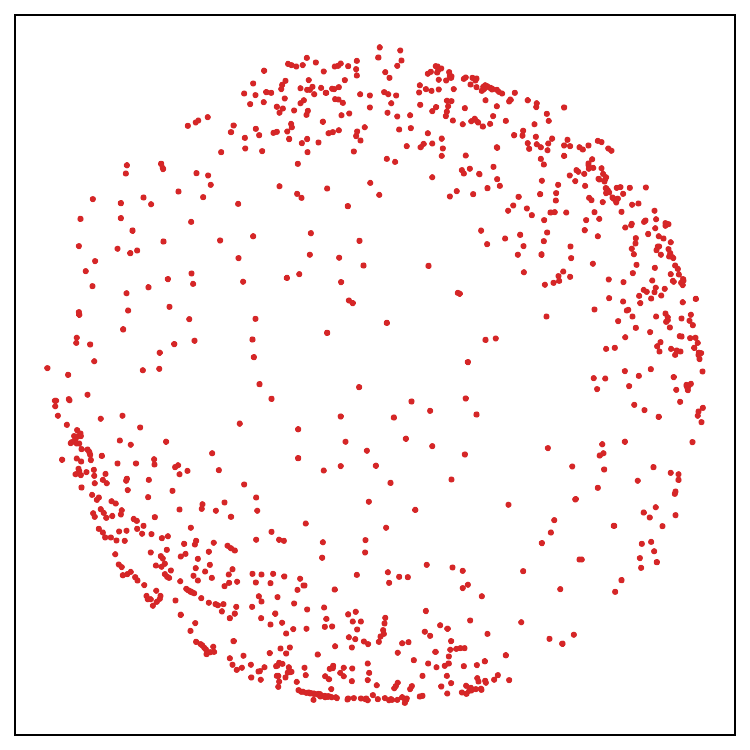}
    \vspace{0pt}
    {\small (a) 4-Qubit 1D\_XYZ}
\end{minipage}%
\hfill
%%%%%%%%%%%%%%%%%%%%%%%%%%%%%%%%
\begin{minipage}[t]{0.48\columnwidth}
    \centering
    \includegraphics[width=0.48\linewidth]{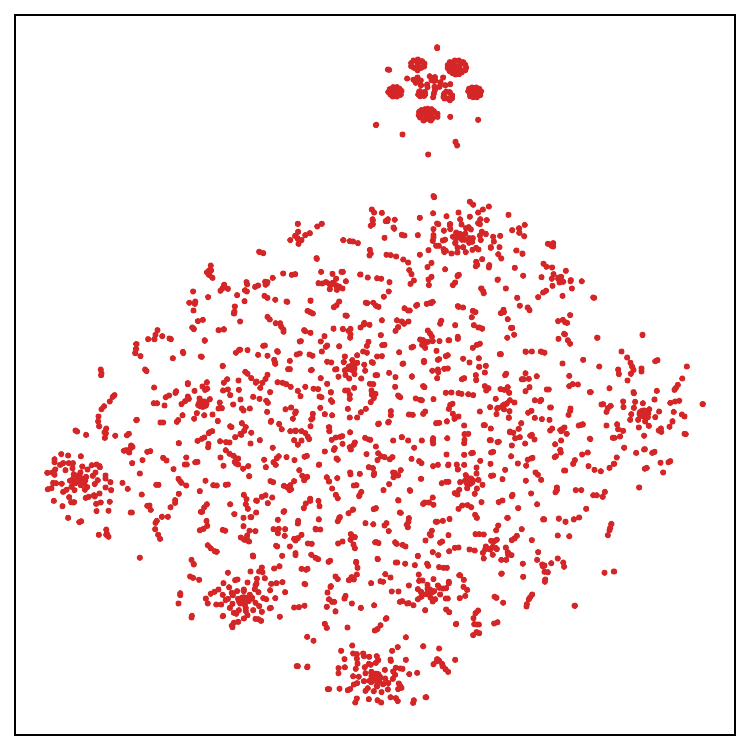}%
    \includegraphics[width=0.48\linewidth]{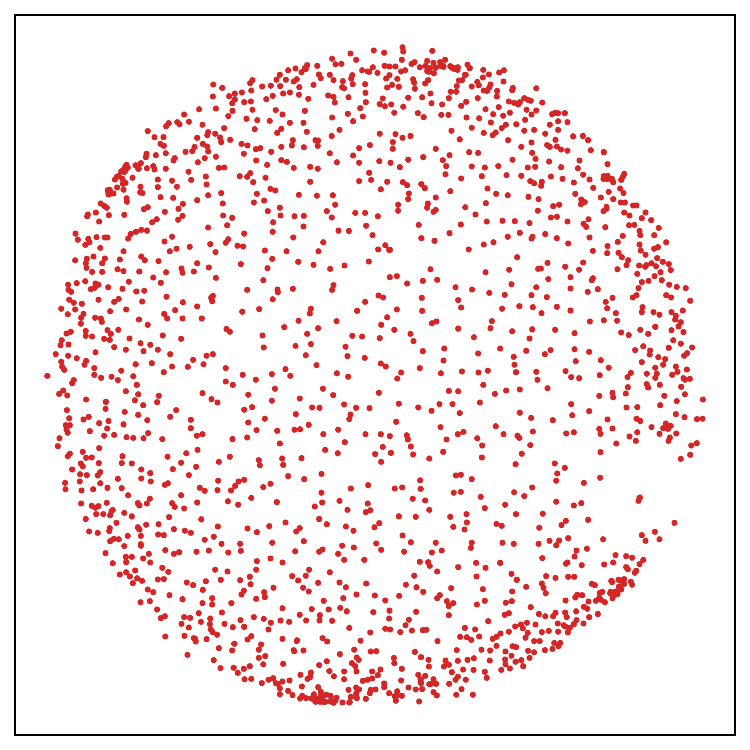}
    \vspace{0pt}
    {\small (b) 12-Qubit 1D\_XYZ}
\end{minipage}

\vspace{-8pt}
\caption{Visualization of optimized parameters using t-SNE (left) and MDS (right) for the 1D Heisenberg XYZ model with circuit implementations of (a) 4 qubits and (b) 12 qubit.}
\label{fig:xyz_tsne_mds}
\vspace{-0.2in}
\end{figure}
%%%%%%%%%%%%%%%%%%%%%%%%%%%%%%%%%%%%%%%%%%%%%%%%%%%%%%%%%%%%%%%%%%%%%

\subsection{Problem Hamiltonian Generation}

The~\design~dataset encompasses three major VQE applications: quantum many-body physics, quantum chemistry, and random VQE. 
% The first two represent the primary scientific domains—physical and molecular systems—where VQE is broadly applicable, with three representative tasks constructed for each. The third domain, random VQE, serves as a widely used benchmark for algorithm evaluation. 
For each domain, coefficients are sampled to generate diverse VQE instances using \texttt{PennyLane}~\cite{bergholmPennyLaneAutomaticDifferentiation2022} and \texttt{TorchQuantum}~\cite{hanruiwang2022quantumnas}. 
Sampled coefficients are stored in \texttt{.h5} files %(available at \TODO{url}), 
and the corresponding Hamiltonian structures are provided in the repository’s Python scripts. 
The sampling procedure is detailed below.

\textbf{Quantum Many-Body Physics}.
We consider three tasks in this domain: the one-dimensional Heisenberg XYZ (1D\_XYZ) model, the one-dimensional Fermi–Hubbard (1D\_FH) model, and the two-dimensional Transverse-Field Ising (2D\_TFI) model. 
All Hamiltonians follow the definitions implemented in the \texttt{PennyLane} library~\cite{bergholmPennyLaneAutomaticDifferentiation2022}. 

For 1D\_XYZ, we use the Hamiltonian in Equation~\ref{eq:heisenberg_xyz} on 4- and 12-qubit chains without boundary conditions. 
The parameters $(J_1, J_2, J_3)$ represent the spin–spin coupling constants along the $X$, $Y$, and $Z$ directions, respectively. 
We sample 2000 unique tuples $(J_1, J_2, J_3) \in [-3, 3]\times[-3, 3]\times[-3, 3]$ with a spacing of $0.1$ for both the 4- and 12-qubit cases. 

For 1D\_FH, we consider spin chains of length 4, 6, and 8, with the Hamiltonian defined in Equation~\ref{eq:1d_fh}. 
Here, $t$ denotes the electron hopping strength between adjacent sites, and $U$ is the on-site Coulomb interaction. 
For each chain length, we sample 1000 unique parameter pairs $(t, U) \in [0, 5]\times[0, 5]$ with a spacing of $0.1$, yielding 3000 instances in total. 

For 2D\_TFI, we use the Hamiltonian in Equation~\ref{eq:2d_tfi} on a $2\times4$ lattice with nearest-neighbor coupling. 
In this model, $j$ is the spin–spin coupling strength, and $\mu$ is the transverse magnetic field. 
We sample 1000 unique parameter pairs $(j, \mu) \in [0, 5]\times[0, 5]$ with a spacing of $0.1$. 
\setlength{\abovedisplayskip}{4pt}
\setlength{\belowdisplayskip}{4pt}
\setlength{\abovedisplayshortskip}{0pt}
\setlength{\belowdisplayshortskip}{0pt}
\begin{align}
\label{eq:heisenberg_xyz}
    \hat{H} &= \sum_{i=0}^3(J_1 X_i X_{i+1} + J_2 Y_i Y_{i+1} + J_3 Z_i Z_{i+1}), \\
\label{eq:1d_fh}
    \hat{H} &= -t\sum_{i=0}^{n-1} (\hat{c}^\dagger_i \hat{c}_{i+1} + \hat{c}_{i+1}\hat{c}) + U\sum_{i=0}^{n-1}\hat{n}_i \hat{n}_{i+1},\\
\label{eq:2d_tfi}
    \hat{H} &= -j \sum_{i, j}Z_i Z_j - \mu \sum_{i} Z_i.
\end{align}

\textbf{Quantum Chemistry}.
We include three molecular Hamiltonians: \ce{H2}, \ce{HeH+}, and \ce{NH3}. 
The dataset contains 150 configurations of \ce{H2}, 1000 of \ce{HeH+}, and 160 of \ce{NH3}. 
Configurations are generated by varying the corresponding bond lengths. 
The Hamiltonians, expressed in Pauli matrix form, are obtained from~\cite{Utkarsh2023Chemistry}, and all molecular Hamiltonians are represented in the STO-3G basis. 
Specifically, for \ce{H2}, we generate 150 Hamiltonians by varying the H–H bond length from 0.5~\AA{} to 4.97~\AA{} in increments of 0.03~\AA{}. 
For \ce{HeH+}, we generate 1000 Hamiltonians by uniformly sampling the He–H bond length from 0.5~\AA{} to 20~\AA{}. 
For \ce{NH3}, we generate 160 Hamiltonians by varying the N–H bond length from 0.5~\AA{} to 2.6~\AA{} in increments of 0.01~\AA{}.

\textbf{Random VQE}.
Randomly generated Hamiltonians are widely used as benchmarks~\cite{liQASMBenchLowLevelQuantum2023} for evaluating VQE algorithms. 
Unlike Hamiltonians derived from specific physical or molecular systems, their inherent randomness mitigates structural bias and provides a diverse, unbiased testing environment. 
Following the procedure in~\cite{liQASMBenchLowLevelQuantum2023}, we construct the Random VQE dataset in~\design~by generating 2800 four-qubit Hamiltonians with random half-integer Pauli string coefficients sampled from the range $[0, 2.5]$.

\vspace{-10pt}
\subsection{Ansatz Circuit Selection}
\vspace{-6pt}

The choice of ansatz is a critical design factor for the performance of VQEs~\cite{tilly2022variational}. 
Following established common practice~\cite{leeQMAMLQuantumModelAgnostic2025a, schuldCircuitcentricQuantumClassifiers2020}, we adopt different ansatz structures across application domains to balance expressivity and trainability.  
For quantum many-body physics, we use the \texttt{CZRXRY} ansatz~\cite{leeQMAMLQuantumModelAgnostic2025a, zhangEscapingBarrenPlateau2022} (Fig.~\ref{fig:ansatz}(a)), which alternates controlled-\texttt{Z} entangling gates with single-qubit \texttt{RX} and \texttt{RY} rotations, providing efficient entanglement while mitigating barren plateaus.  
For molecular Hamiltonians, we adopt the strongly entangling ansatz (Fig.~\ref{fig:ansatz}(b)), widely employed in quantum chemistry studies~\cite{leeQMAMLQuantumModelAgnostic2025a, schuldCircuitcentricQuantumClassifiers2020}. 
This ansatz offers sufficient expressivity to capture electronic structure while remaining scalable across molecular systems.  
For random VQE benchmarking, we employ the \texttt{U3CU3} ansatz~\cite{hanruiwang2022quantumnas} (Fig.~\ref{fig:ansatz}(c)), which combines \texttt{U3} single-qubit rotations with controlled-\texttt{U3} entangling gates, offering flexibility and serving as a widely adopted benchmark for variational algorithm evaluations.

\vspace{-6pt}
\subsection{VQE Optimization}
\vspace{-4pt}
The choice of optimizer has a critical impact on the convergence and overall performance of VQEs~\cite{tilly2022variational}. 
Prior studies highlight trade-offs between gradient-free methods such as \texttt{COBYLA}, which can be effective for small-scale problems but scale poorly, and gradient-based methods such as \texttt{SLSQP} and \texttt{Adam}, which leverage gradient information or approximations to improve efficiency. 
Considering the balance between optimization performance, computational cost, and GPU acceleration support, we adopt \texttt{Adam} with a learning rate of $10^{-3}$ as the optimizer for all applications in~\design.  
All experiments were conducted on an AMD Ryzen 5 1600 CPU with acceleration from an NVIDIA RTX 3090 GPU. 
Dataset collection required more than 200 hours of wall-clock time.

\begin{comment}
    
\begin{table*}[t!]\centering
    \caption{Optimization settings in \design.}
    \label{tab:opt_settings}
    \setlength{\tabcolsep}{1.8pt}
    \begin{tabular}{c c c c c c}
    \hline \hline
    Domain & Application &Optimization step & Learning rate & Backend & Ansatz\\
    \hline \hline
    \multirow{3}{*}{Quantum Many-Body Physics} & 1D Heisenberg XYZ &\multirow{3}{*}{$2000$} & \multirow{3}{*}{$10^{-2}$} & \multirow{3}{*}{\texttt{PennyLane}} & \multirow{3}{*}{as in Fig.~\ref{fig:ansatz}(a)}\\
    & 2D TFI & & & & \\
    & 1D FH & & & & \\
    \hline
    \multirow{3}{*}{Quantum Chemistry} & \ce{H2} &\multirow{3}{*}{$2000$} & \multirow{3}{*}{$10^{-3}$} & \multirow{3}{*}{\texttt{PennyLane}} & \multirow{3}{*}{as in Fig.~\ref{fig:ansatz}(b)}\\
    & \ce{HeH+} & & & & \\
    & \ce{NH3} & & & & \\
    \hline
    \multirow{1}{*}{NISQ Device Benchmarking} & Random VQE & $500$ & $5 \times 10^{-3}$ & \texttt{TorchQuantum} & as in Fig.~\ref{fig:ansatz}(c) \\
    \hline \hline
    \end{tabular}
\end{table*}

\end{comment}

\begin{figure*}[t]
\centering
\makebox[\textwidth][c]{%
\begin{subfigure}[t]{0.14\textwidth}
    \centering
    \includegraphics[width=\linewidth]{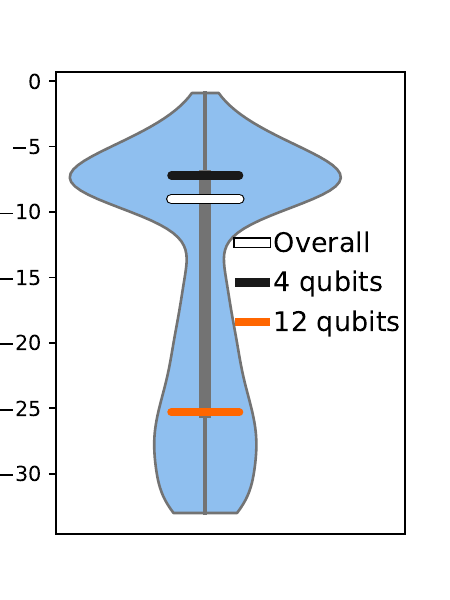}
    \vspace{-0.3in}
    \caption{1D\_XYZ}
    \label{fig:1d_heisenberg_xyz_violin}
\end{subfigure}
\hspace{-0.4em}
\begin{subfigure}[t]{0.14\textwidth}
    \centering
    \includegraphics[width=\linewidth]{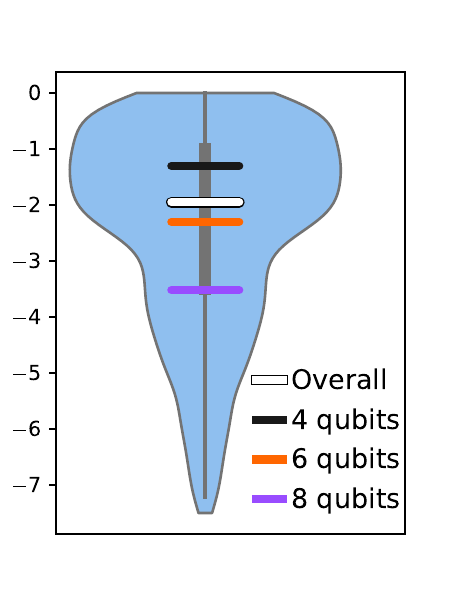}
    \vspace{-0.3in}
    \caption{1D\_FH}
    \label{fig:1d_fh_violin}
\end{subfigure}
\hspace{-0.4em}
\begin{subfigure}[t]{0.14\textwidth}
    \centering
    \includegraphics[width=\linewidth]{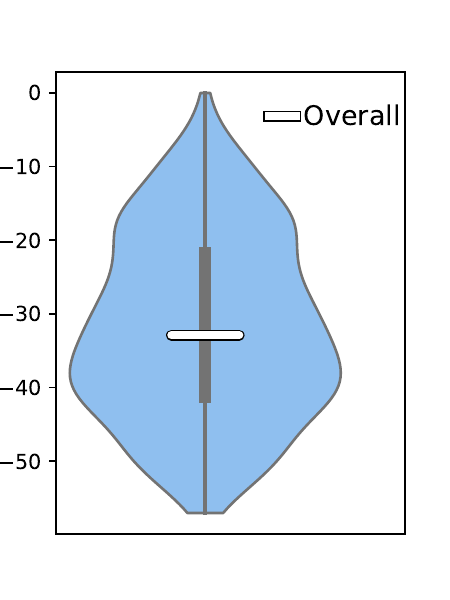}
    \vspace{-0.3in}
    \caption{2D\_TFI}
    \label{fig:2d_tfi_violin}
\end{subfigure}
\hspace{-0.4em}
\begin{subfigure}[t]{0.14\textwidth}
    \centering
    \includegraphics[width=\linewidth]{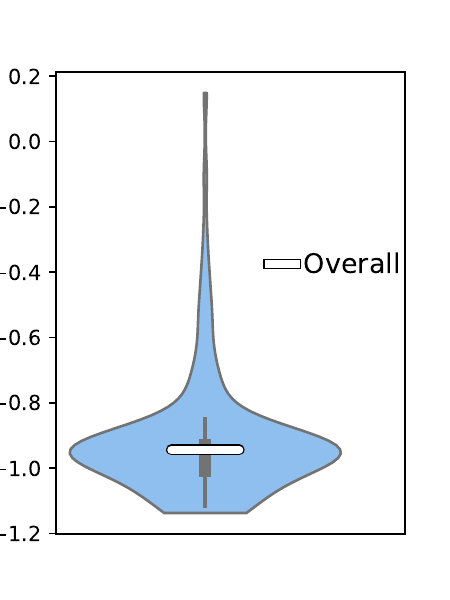}
    \vspace{-0.3in}
    \caption{\ce{H2}}
    \label{fig:h2_violin}
\end{subfigure}
\hspace{-0.4em}
\begin{subfigure}[t]{0.14\textwidth}
    \centering
    \includegraphics[width=\linewidth]{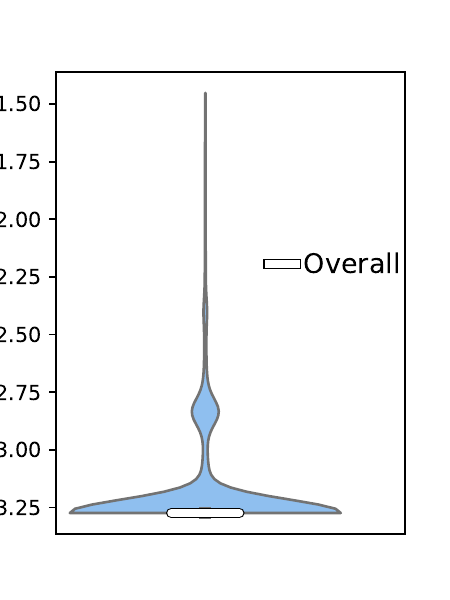}
    \vspace{-0.3in}
    \caption{\ce{HeH+}}
    \label{fig:hehp_violin}
\end{subfigure}
\hspace{-0.4em}
\begin{subfigure}[t]{0.14\textwidth}
    \centering
    \includegraphics[width=\linewidth]{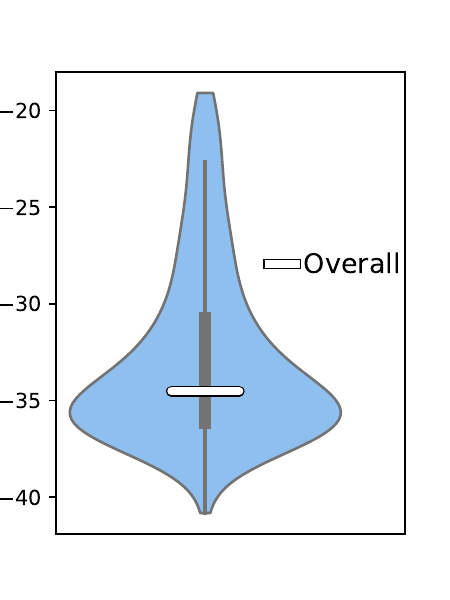}
    \vspace{-0.3in}
    \caption{\ce{NH3}}
    \label{fig:nh3_violin}
\end{subfigure}
\hspace{-0.4em}
\begin{subfigure}[t]{0.14\textwidth}
    \centering
    \includegraphics[width=\linewidth]{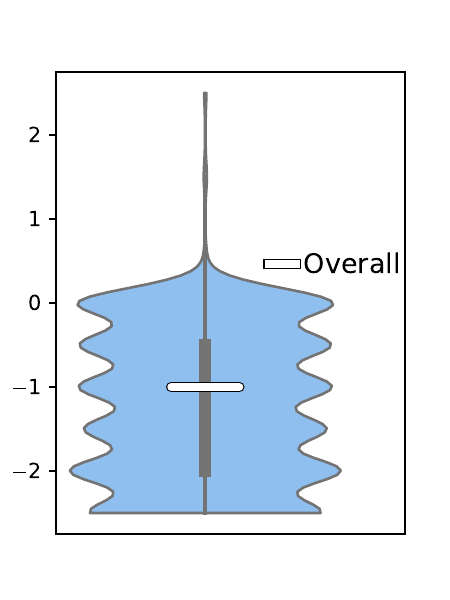}
    \vspace{-0.3in}
    \caption{Random\_VQE}
    \label{fig:random_vqe_violin}
\end{subfigure}
}
\vspace{-0.2in}
\caption{Violin plots of optimized ground-state energies in~\design. Each plot shows the distribution of ground-state energies for individual tasks across different application domains.}
\label{fig:violin_plots}
\vspace{-0.23in}
\end{figure*}
%%%%%%%%%%%%%%%%%%%%%%%%%%%%%%%%%%%%%%%%%%%%%%%%%%%%%%%%%%%%%%%%%%%%%

\vspace{-12pt}
\section{Data Characterization}
\vspace{-8pt}
We characterize the \design~dataset by analyzing and visualizing optimized VQE parameter distributions.  
We apply two common dimensionality reduction techniques—t-distributed Stochastic Neighbor Embedding (\texttt{t-SNE}) and Multidimensional Scaling (MDS)—to project the high-dimensional space into two dimensions.  
\texttt{t-SNE} preserves local neighborhoods and highlights clustering, while MDS emphasizes global pairwise distances, revealing overall geometry.

\textbf{VQE Parameter Distributions across Applications.}  
As illustrated in Fig.~\ref{fig:domain_tsne_mds}, optimized parameters from different tasks form well-defined clusters. In the quantum many-body domain (Fig.~\ref{fig:domain_tsne_mds}a), tasks such as 1D\_XYZ, 1D\_FH, and 2D\_TFI exhibit clearly separated parameter distributions. In the quantum chemistry domain (Fig.~\ref{fig:domain_tsne_mds}b), molecular systems including \ce{H2}, \ce{HeH+}, and \ce{NH3} are distinctly partitioned according to bond-length variations. By contrast, the random VQE benchmarking domain (Fig.~\ref{fig:domain_tsne_mds}c) presents a more diffuse structure, consistent with its stochastic construction.
Together, these visualizations demonstrate that optimized VQE parameters not only capture the underlying physics and chemistry of the problem Hamiltonians but also retain sufficient diversity to distinguish domains and tasks.  
This underscores the utility of~\design~as a resource for analyzing parameter landscapes and investigating transferability across VQE applications.

\textbf{Symmetry in Optimized Parameters for 1D\_FH.}  
In Fig.~\ref{fig:fh_tsne_mds}, we show the optimized parameter distributions of 1D\_FH for 4, 6, and 8 qubits.
The distributions exhibit clear symmetries, similar to the QAOA case~\cite{akshay2021parameter}.  
These symmetries reflect the underlying geometry of the Hilbert space of the 1D\_FH model.  
As the number of qubits increases, additional symmetries emerge—for example, the 8-qubit case in Fig.~\ref{fig:fh_tsne_mds}(c) exhibits the richer $O(2)$ symmetry compared to the $C_4$ symmetry observed in Fig.~\ref{fig:fh_tsne_mds}(a)—resulting in a more complex landscape of optimized parameters.

\textbf{Clustering in Optimized Parameters for 1D\_XYZ.}  
In Fig.~\ref{fig:xyz_tsne_mds}, we plot the distribution of optimized parameters for 1D\_XYZ with different qubit sizes (4 and 12 qubits).  
For the 4-qubit case, distinct clustering is observed, reflecting the internal symmetries of the 1D\_XYZ model.  
As the number of qubits increases, the clustering becomes less pronounced but still persists, suggesting that the internal symmetries remain influential even in larger systems.  
To more clearly reveal these symmetries at larger scales, additional data points are needed to better capture the underlying structure.

\textbf{Distributions of Optimized Ground-State Energies.}  
Fig.~\ref{fig:violin_plots} visualizes the distributions of optimized ground-state energies across different tasks. Each violin plot represents the energy distribution within an application domain, with the width indicating the density of data points at each level.
In the quantum many-body physics domain (Fig.~\ref{fig:violin_plots}(a)$\sim$(c)), the major modes of the distributions align with the number of qubits, as reflected in the medians coinciding with the peaks of the violins.  
The quantum chemistry domain (Fig.~\ref{fig:violin_plots}(d)$\sim$(f)) exhibits a broader spread of ground-state energies but still reveals modes associated with qubit counts.  
By contrast, the random VQE application (Fig.~\ref{fig:violin_plots}(g)) shows modes that are independent of qubit number—all instances being 4-qubit—and instead arise from the discrete Hamiltonian structure.  
Because the ground-state energies are determined by the coefficients of Pauli strings, they exhibit discretized energy gaps, which manifest as multiple modes in the violin plots.

\vspace{-24pt}
\section{Potential Applications}
\label{sec:app}
\vspace{-6pt}

This section highlights several research directions in which the~\design~dataset can provide substantial benefits.

\noindent{\bf VQE Initialization and Optimization}.  
\design~supports VQE parameter initialization by providing starting points that lower initial loss, accelerate convergence, and improve final performance. It further extends to recent machine-learning–based initialization strategies across diverse domains~\cite{mesman2024nn, Miao:PRA2024, zhang2025diffusion}. Unlike prior small and sparsely sampled datasets lacking optimization trajectories, \design~captures ground-state energy evolution, parameter dynamics, and barren plateau behavior, making it a valuable resource for both practical optimization and theoretical studies of VQEs.

\noindent{\bf Transfer Learning across VQE Tasks}.  
\design~spans three major VQE application domains, offering seven tasks with thousands of data points each—far exceeding the limited hundreds in prior datasets. This scale enables systematic studies of parameter transferability~\cite{zhuangComprehensiveSurveyTransfer2021} and model-agnostic meta-learning~\cite{finnModelAgnosticMetaLearningFast2017}, which were previously constrained by data scarcity. Recent work~\cite{leeQMAMLQuantumModelAgnostic2025a} demonstrates that meta-trained neural networks can generalize across Hamiltonians, mitigate barren plateaus, and accelerate convergence. The task diversity in \design~also supports few-shot adaptation methods that reduce optimization overhead for unseen quantum systems.

\noindent{\bf VQE Architecture Design and Search}.  
\design~serves as a benchmark for architecture exploration in VQE circuits. Recent frameworks~\cite{hanruiwang2022quantumnas, wuQuantumDARTSDifferentiableQuantum2023,furrutterQuantumCircuitSynthesis2024b} leverage deep learning to search ansatz architectures and parameters, showing promise for robust design in the NISQ era. The curated parameters and ground-state energies in \design~further enable models that generate task-specific ansatz architectures.

\vspace{-6pt}
\section{Conclusion and Future Work}
\label{sec:conc}
\vspace{-4pt}

We open-source the \design~dataset, which currently spans three major VQE application domains with 12,110 instances. 
Future work includes:  
(1) Expanding coverage to additional quantum chemistry tasks, particularly larger molecules such as \ce{C2H6} and \ce{CH4}, to build a comprehensive NISQ-era VQE dataset.  
(2) Enriching labels with figures of merit (FoMs) such as fidelity, depth, and execution time to better support VQE architecture search.  
(3) Incorporating hardware data by collecting optimization trajectories on IBM and IonQ devices to improve practical relevance.

% \noindent{\bf  ACKNOWLEDGMENT}

%\newpage

\clearpage

\let\oldbibliography\thebibliography
\renewcommand{\thebibliography}[1]{\oldbibliography{#1}
\setlength{\itemsep}{-3pt}}

\bibliographystyle{IEEEbib.bst}
%{\small
\bibliography{refs.bib}

\begin{thebibliography}{10}

\bibitem{peruzzo2014variational}
Alberto Peruzzo, Jarrod McClean, Peter Shadbolt, Man-Hong Yung, Xiao-Qi Zhou, Peter~J Love, Al{\'a}n Aspuru-Guzik, and Jeremy~L O’brien,
\newblock ``A variational eigenvalue solver on a photonic quantum processor,''
\newblock {\em Nature communications}, vol. 5, no. 1, pp. 4213, 2014.

\bibitem{mcclean2016theory}
Jarrod~R McClean, Jonathan Romero, Ryan Babbush, and Al{\'a}n Aspuru-Guzik,
\newblock ``The theory of variational hybrid quantum-classical algorithms,''
\newblock {\em New Journal of Physics}, vol. 18, no. 2, pp. 023023, 2016.

\bibitem{bharti2022noisy}
Kishor Bharti, Alba Cervera-Lierta, Thi~Ha Kyaw, Tobias Haug, Sumner Alperin-Lea, Abhinav Anand, Matthias Degroote, Hermanni Heimonen, Jakob~S Kottmann, Tim Menke, et~al.,
\newblock ``Noisy intermediate-scale quantum algorithms,''
\newblock {\em Reviews of Modern Physics}, vol. 94, no. 1, pp. 015004, 2022.

\bibitem{bauer2020quantum}
Bela Bauer, Sergey Bravyi, Mario Motta, and Garnet Kin-Lic Chan,
\newblock ``Quantum algorithms for quantum chemistry and quantum materials science,''
\newblock {\em Chemical reviews}, vol. 120, no. 22, pp. 12685--12717, 2020.

\bibitem{hensgens2017quantum}
Toivo Hensgens, Takafumi Fujita, Laurens Janssen, Xiao Li, CJ~Van~Diepen, Christian Reichl, Werner Wegscheider, Sankar Das~Sarma, and Lieven~MK Vandersypen,
\newblock ``Quantum simulation of a fermi--hubbard model using a semiconductor quantum dot array,''
\newblock {\em Nature}, vol. 548, no. 7665, pp. 70--73, 2017.

\bibitem{li2019variational}
Yifan Li, Jiaqi Hu, Xiao-Ming Zhang, Zhigang Song, and Man-Hong Yung,
\newblock ``Variational quantum simulation for quantum chemistry,''
\newblock {\em Advanced Theory and Simulations}, vol. 2, no. 4, pp. 1800182, 2019.

\bibitem{tilly2022variational}
Jules Tilly, Hongxiang Chen, Shuxiang Cao, Dario Picozzi, Kanav Setia, Ying Li, Edward Grant, Leonard Wossnig, Ivan Rungger, George~H Booth, et~al.,
\newblock ``The variational quantum eigensolver: a review of methods and best practices,''
\newblock {\em Physics Reports}, vol. 986, pp. 1--128, 2022.

\bibitem{mesman2024nn}
Koen Mesman, Yinglu Tang, Matthias Moller, Boyang Chen, and Sebastian Feld,
\newblock ``Nn-ae-vqe: Neural network parameter prediction on autoencoded variational quantum eigensolvers,''
\newblock {\em arXiv preprint arXiv:2411.15667}, 2024.

\bibitem{Miao:PRA2024}
Jiaqi Miao, Chang-Yu Hsieh, and Shi-Xin Zhang,
\newblock ``Neural-network-encoded variational quantum algorithms,''
\newblock {\em Physical Review Applied}, vol. 21, pp. 014053, Jan 2024.

\bibitem{zhang2025diffusion}
Shikun Zhang, Zheng Qin, Yongyou Zhang, Yang Zhou, Rui Li, Chunxiao Du, and Zhisong Xiao,
\newblock ``Diffusion-enhanced optimization of variational quantum eigensolver for general hamiltonians,''
\newblock {\em arXiv preprint arXiv:2501.05666}, 2025.

\bibitem{liQASMBenchLowLevelQuantum2023}
Ang Li, Samuel Stein, Sriram Krishnamoorthy, and James Ang,
\newblock ``{{QASMBench}}: {{A Low-Level Quantum Benchmark Suite}} for {{NISQ Evaluation}} and {{Simulation}},''
\newblock {\em ACM Transactions on Quantum Computing}, vol. 4, no. 2, pp. 10:1--10:26, Feb. 2023.

\bibitem{nakayama2023vqegenerated}
Akimoto Nakayama, Kosuke Mitarai, Leonardo Placidi, Takanori Sugimoto, and Keisuke Fujii,
\newblock ``Vqe-generated quantum circuit dataset for machine learning,'' 2023.

\bibitem{leeQMAMLQuantumModelAgnostic2025a}
Junyong Lee, JeiHee Cho, and Shiho Kim,
\newblock ``Q-maml: Quantum model-agnostic meta-learning for variational quantum algorithms,'' 2025.

\bibitem{Utkarsh2023Chemistry}
Utkarsh Azad,
\newblock ``Pennylane quantum chemistry datasets,'' \url{https://pennylane.ai/datasets/collection/qchem}, 2023.

\bibitem{zhangEscapingBarrenPlateau2022}
Kaining Zhang, Liu Liu, Min-Hsiu Hsieh, and Dacheng Tao,
\newblock ``Escaping from the barren plateau via {{Gaussian}} initializations in deep variational quantum circuits,''
\newblock in {\em Proceedings of the 36th {{International Conference}} on {{Neural Information Processing Systems}}}, 2022, pp. 18612--18627.

\bibitem{schuldCircuitcentricQuantumClassifiers2020}
Maria Schuld, Alex Bocharov, Krysta~M. Svore, and Nathan Wiebe,
\newblock ``Circuit-centric quantum classifiers,''
\newblock {\em Physical Review A}, vol. 101, no. 3, pp. 032308, Mar. 2020.

\bibitem{hanruiwang2022quantumnas}
Hanrui Wang, Yongshan Ding, Jiaqi Gu, Zirui Li, Yujun Lin, David~Z Pan, Frederic~T Chong, and Song Han,
\newblock ``Quantumnas: Noise-adaptive search for robust quantum circuits,''
\newblock in {\em The 28th IEEE International Symposium on High-Performance Computer Architecture (HPCA-28)}, 2022.

\bibitem{bergholmPennyLaneAutomaticDifferentiation2022}
Ville Bergholm et~al.,
\newblock ``{{PennyLane}}: {{Automatic}} differentiation of hybrid quantum-classical computations,'' July 2022.

\bibitem{akshay2021parameter}
Vishwanathan Akshay, Daniil Rabinovich, Ernesto Campos, and Jacob Biamonte,
\newblock ``Parameter concentrations in quantum approximate optimization,''
\newblock {\em Physical Review A}, vol. 104, no. 1, pp. L010401, 2021.

\bibitem{zhuangComprehensiveSurveyTransfer2021}
Fuzhen Zhuang, Zhiyuan Qi, Keyu Duan, Dongbo Xi, Yongchun Zhu, Hengshu Zhu, Hui Xiong, and Qing He,
\newblock ``A {{Comprehensive Survey}} on {{Transfer Learning}},''
\newblock {\em Proceedings of the IEEE}, vol. 109, no. 1, pp. 43--76, Jan. 2021.

\bibitem{finnModelAgnosticMetaLearningFast2017}
Chelsea Finn, Pieter Abbeel, and Sergey Levine,
\newblock ``Model-{{Agnostic Meta-Learning}} for {{Fast Adaptation}} of {{Deep Networks}},''
\newblock in {\em Proceedings of the 34th {{International Conference}} on {{Machine Learning}}}. July 2017, pp. 1126--1135, PMLR.

\bibitem{wuQuantumDARTSDifferentiableQuantum2023}
Wenjie Wu, Ge~Yan, Xudong Lu, Kaisen Pan, and Junchi Yan,
\newblock ``{{QuantumDARTS}}: {{Differentiable Quantum Architecture Search}} for {{Variational Quantum Algorithms}},''
\newblock in {\em Proceedings of the 40th {{International Conference}} on {{Machine Learning}}}. July 2023, pp. 37745--37764, PMLR.

\bibitem{furrutterQuantumCircuitSynthesis2024b}
Florian F{\"u}rrutter, Gorka {Mu{\~n}oz-Gil}, and Hans~J. Briegel,
\newblock ``Quantum circuit synthesis with diffusion models,''
\newblock {\em Nature Machine Intelligence}, vol. 6, no. 5, pp. 515--524, May 2024.

\end{thebibliography}
%}

\end{document}